\documentclass[10pt,twocolumn,letterpaper]{article}

\usepackage{iccv}
\usepackage{times}
\usepackage{epsfig}
\usepackage{graphicx}
\usepackage{amsmath}
\usepackage{amssymb}

\usepackage{subfigure}
\usepackage{xspace}
\usepackage[footnotesize,bf]{caption}
\usepackage{setspace}
\usepackage{dsfont}
\usepackage{algorithmic}
\usepackage{algorithm}

\def\etal{{et al.}\xspace}

\newcommand\norm[1]{\left\lVert#1\right\rVert}

\newcommand\degsymb[0]{$^\circ$}

\usepackage[pagebackref=true,breaklinks=true,letterpaper=true,colorlinks,bookmarks=false]{hyperref}

 \iccvfinalcopy 

\def\iccvPaperID{7} 

\ificcvfinal\fi
\begin{document}
\raggedbottom

\title{A-MAL: Automatic Movement Assessment Learning from Properly Performed Movements in 3D Skeleton Videos}

\author{Tal Hakim and Ilan Shimshoni\\
The University of Haifa, Haifa, Israel\\
{\tt\small thakim.research@gmail.com}, {\tt\small ishimshoni@mis.haifa.ac.il}}

\maketitle
\thispagestyle{empty}


\begin{abstract}
The task of assessing movement quality has recently gained high demand in a variety of domains. The ability to automatically assess subject movement in videos that were captured by affordable devices, such as Kinect cameras, is essential for monitoring clinical rehabilitation processes, for improving motor skills and for movement learning tasks. The need to pay attention to low-level details while accurately tracking the movement stages, makes this task very challenging. In this work, we introduce A-MAL \ificcvfinal\footnote{Code available at: \url{http://github.com/skvp-owner/a-mal}}\fi, an automatic, strong movement assessment learning algorithm that only learns from properly-performed movement videos without further annotations, powered by a deviation time-segmentation algorithm, a parameter relevance detection algorithm, a novel time-warping algorithm that is based on automatic detection of common temporal points-of-interest and a textual-feedback generation mechanism. We demonstrate our method on movements from the Fugl-Meyer Assessment (FMA) test, which is typically held by occupational therapists in order to monitor patients' recovery processes after strokes.

\end{abstract}

\section{Introduction}
The capability of automatic movement assessment has recently gained high demand. One of the most significant reasons is the availability of low-cost 3D skeleton recognition devices, such as Kinect, which redefine the target audience of applications that are based on user pose and movement, including applications for movement assessment or movement learning, as well as other tasks, such as surveillance, entertainment and exercise. Addressing this problem is considered a hard task, especially when compared to the other researched tasks in the 3D skeleton-video domain, which usually give weaker attention to timings, performances and low-level measurements.

\begin{figure}[]
\centering
  \includegraphics[width=\linewidth,keepaspectratio]{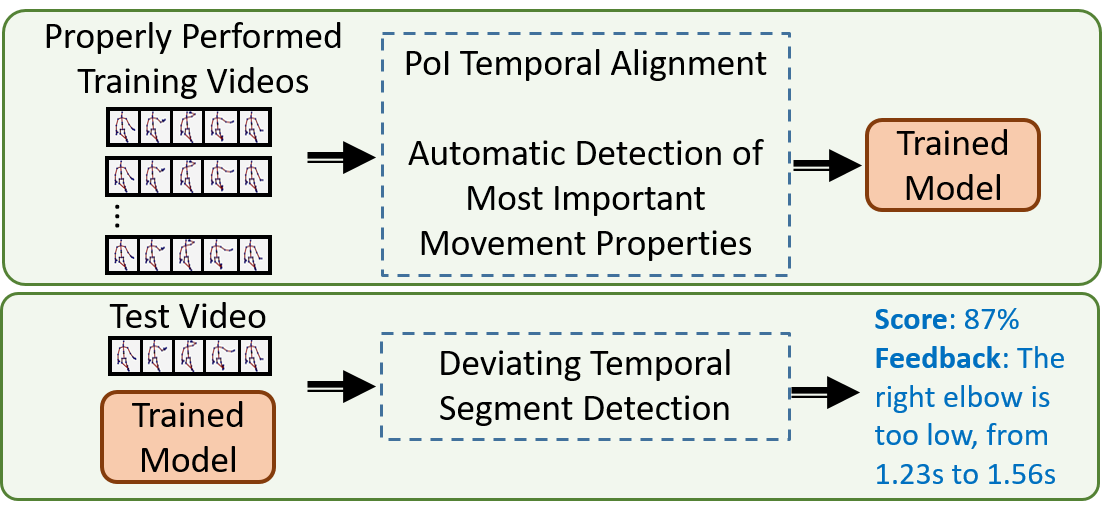}
  \caption[The proposed system]{The main principles of the proposed system. A model is learned solely from properly performed movements and without further annotations, by automatic extraction of movement properties, temporal alignment and learning of a frame-level statistical model. The model is used to produce score and feedback by detecting deviating temporal segments.}
  \label{fig:FrontFigure}
\end{figure}

The task of automatic movement assessment is very essential for medical needs. One outstanding example is the Fugl-Meyer Assessment (FMA)~\cite{fma} test, which is typically held by occupational therapists and has a numerical score. The test is performed from time to time, mainly on people who are recovering after strokes and aims to monitor their recovery processes. Except for the purpose of enabling FMA tests to be held at home, without the presence of an occupational therapist, the automation of this test has the major benefit of increased reliability, by discarding the error caused by the fact that a patient is sometimes assessed by different people at different times.

In this work, we aim to provide a solution for the general movement assessment task. Given a training set of 3D skeleton videos of a properly-performed type of movement, performed by different people, our goal is to build a statistical model that will produce feedback and a score and will deal with the challenges of the assessment task, such as identifying and paying attention to the most important low-level features and tracking the movement stages. Our resulting model produces numerical scores that accurately discriminate between the possible scores of the FMA assessment, as demonstrated in Section~\ref{lbl:results}. The main principles of the proposed system are illustrated in Figure~\ref{fig:FrontFigure}.

We eliminate the joint location differences related to the camera location and the subject skeleton dimensions and then time-align the training videos, by automatically detecting mutual temporal points-of-interest (PoIs) and forcing them to occur at the same time in all the videos, which makes frame-level statistics more effective. The FMA movements in Figure~\ref{fig:motionRepFrames}, which have structured rests that can have different lengths, especially when assessing improperly performed movements, support the idea of using temporal PoIs for time-alignment, rather than existing off-the-shelf algorithms, such as Dynamic Time Warping (DTW)~\cite{dtw}, which gives an equal weight to all the frames. In assessment mode, we time-segment the frame-level deviations, to eliminate false values and produce effective feedback, based on periods in the movement duration that clearly deviate.

The main contributions of this work are as follows. First, we present a novel step-by-step algorithm for training a model that produces scores and feedback, by only learning from a few videos of a properly-performed type of movement, which were captured using a relatively cheap and noisy Kinect2 device. Second, as a part of the flow, we present a novel time-warping algorithm, which is based on detection of mutual temporal PoIs, which in contrast to DTW, produces a continuous and more accurate alignment, by exploiting the rests that exist in the structure of FMA movements. Third, we present a deviation time-segmentation algorithm that deals with noisy deviations and produces effective feedback. We demonstrate the robustness of our model by showing its capability of accurately discriminating between the possible FMA performance categories using its produced numerical scores and by showing its accurate and effective produced feedback. In addition to the scientific contribution, we developed an open-source Python framework for processing skeleton videos, with a new human-readable 3D skeleton video format and a designated 3D player, which we hope will help facilitate the future research in this domain.

\section{Related Work}
\label{lbl:background_and_related_work}
\paragraph{Skeleton Tasks.}  While there are only a few works that try to provide solutions for the problem of movement assessment from skeleton data, the problems of action recognition and person identification have been widely researched in recent years, especially since affordable devices such as Kinect became available. While solutions for the movement assessment problem can be adapted to perform action recognition, solutions that were designed solely for the action recognition problem use aggregative high level features~\cite{mrAngleSimilaritiesHog,mrVelocityHistogram,mrCovarianceHierarchy,mrDisplacements} and pay lower attention to timings, which is not satisfactory for the assessment task. However, while similarly to movement assessment, action recognition solutions try to get rid of movement properties that are person specific, solutions for of person identification~\cite{piIgor,piNotIgor} try to exploit them to discriminate between people.
\paragraph{Skeleton Representations.} Han \etal~\cite{review} divide the skeleton representations in their review into four main types of representations: Displacement-based representations, which can be pairwise or temporal~\cite{disp1,disp2,disp3,disp4,disp5,disp6,disp7,disp8}, orientation-based representations, which can be pairwise or temporal~\cite{orie1,orie2,orie3,orie4,orie5}, representations based on raw-joint positions, which may only apply geometric fixes and dimensionality reduction methods~\cite{raw1,raw2,raw3,raw4,raw5,raw6} and multi-modal representations, which combine any of the three types of representations~\cite{mult1,mult2,mult3}. In our work, we combine displacement-based and orientation-based features for alignment of videos in time, while combining all the three types of features for computing scores and feedback.

\paragraph{Movement Assessment.} The problem of movement assessment from skeletal data has been researched over the last years. Some systems and algorithms were proposed for both numerical assessment and feedback generation. Paiement \etal~\cite{online} demonstrated assessment of gait-on-stairs movements, proposing an online assessment algorithm that is based on Hidden Markov Models (HMM) for dynamics learning and assessment and on an independent statistical model for pose learning and assessment, after reducing dimensionality and filtering noise using Diffusion Maps~\cite{diffusion}. They used normal occurrences of a type of movement, in order to train a designated statistical model. Su~\cite{assistant} demonstrated assessment of shoulder rehabilitation exercises, describing a personal assistant system that used DTW to match video frame indices to each-other and a fuzzy logic approach to produce a score for each joint. Parisi \etal~\cite{recurrent} demonstrated assessment of power lifting movements, introducing the MGWR recurrent neural network, an architecture that is trained to predict the next frame, and presented an algorithm for real time assessment and feedback generation. Eichler \etal~\cite{nadav1,nadav2} demonstrated high performance in classification of two FMA movements into score categories, combining multiple depth cameras and proposing an algorithm for calibration and improvement of skeleton location predictions. Palma \etal~\cite{hmmdtw} conducted experiments, trying to find out the effectiveness of the DTW and HMM solutions for detecting deviations from normality, during the performance of physical therapy activities. None of the existing works used deep learning, mainly because they all had to deal with very small datasets. 

\paragraph{Temporal Alignment.} The DTW algorithm~\cite{assistant}, finds the optimal monotonic increasing index matching. Its main limitation is the fact that it only produces discrete matching between pairs of indices. HMMs~\cite{online}, on the other hand, use sequential training samples to learn the state transition and observation yielding probabilities. Using Viterbi's algorithm~\cite{viterbi}, HMMs predict the most likely sequence of states that would yield a sequence of observations, under the Markovian assumption. Its main disadvantage is the fact that it requires the training data to be annotated with the movement stages. The Correlation Optimized Warping (COW)~\cite{tomasi2004correlation} algorithm finds the optimal time-scaling coefficients of uniformly-divided sequence segments. Its main limitation is the fact that its scales divisions that are based on timings, rather than on detected events.

\section{Training Method}
\label{lbl:our-approach} 
\begin{figure}[]
\centering
  \includegraphics[width=0.78\linewidth,keepaspectratio]{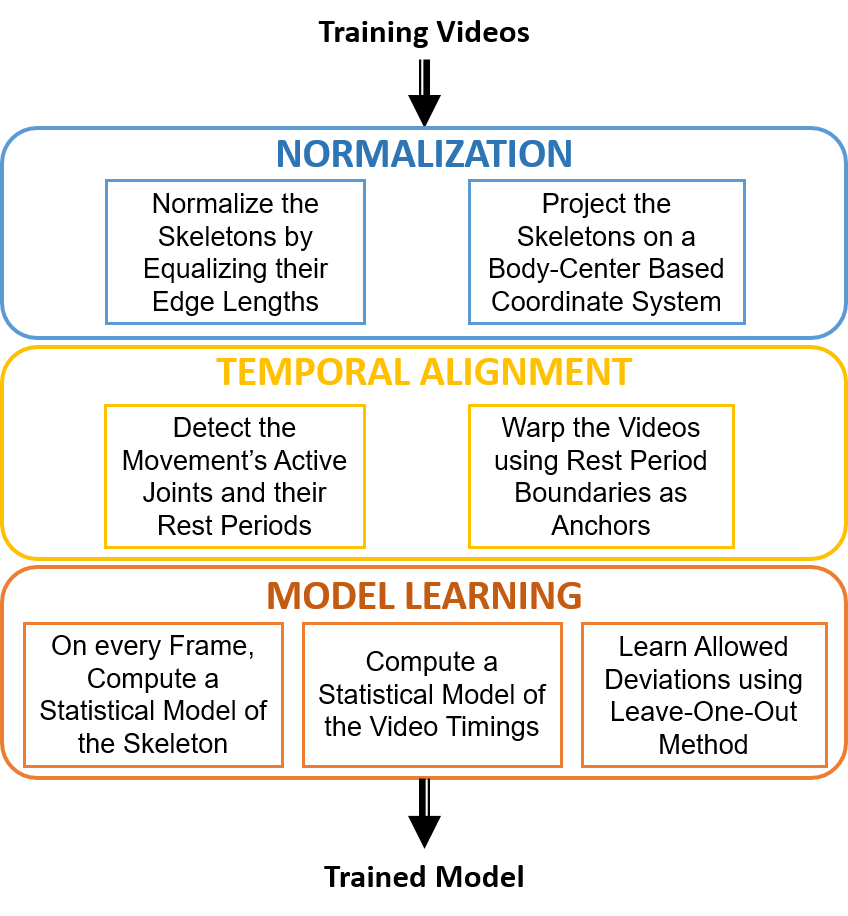}
  \caption[The proposed main learning flow]{The main model learning flow in high level, which can be seen as 3-incremental steps.}
  \label{fig:TrainingMainFlow}
\end{figure}

The method we suggest here, is based on taking a set of videos of different people properly performing the same type of movement and extracting frame-level features. As illustrated in Figure~\ref{fig:TrainingMainFlow}, we normalize all the skeletons and project them on a uniform coordinate system. We then temporally align the videos, such that each movement stage will occur at the same time in all the videos. Having these conditions satisfied, we extract frame-level features and learn a model. In Section~\ref{lbl:our-method-test}, we describe the analysis process that is applied on test videos.
\paragraph{Input Format and Mathematical Notation.}
\label{lbl:mathNotation}
We assume that the input to our algorithm is a list of $n$ skeleton videos, where a skeleton is defined by a set of joints $J$ and the edges connecting them $e = (j_{1}, j_{2}) \in E , j_{1}, j_{2} \in J$. The location $(x, y, z)$ of each joint $j \in J$ is known at each frame $f \in F$ in each video. Let us denote $F_{i}$ as the list of frames in video $i$, and let us denote $L_{jfi}$ as the location of joint $j$ in frame $f \in F_{i}$.

When we use the notation $\widehat{x}$, we refer to the normalized version of vector $\overrightarrow{x}$.
\subsection{Video Normalization}
\label{lbl:videoNormalization}
The video normalization step is essential for making the input videos comparable, by removing unneeded, distracting information from the data.
\paragraph{Skeleton Dimension Normalization.}
\label{lbl:skeleton-dimension-normalization}
We eliminate the dimensional body differences between the people in the training set, by forcing equal skeleton edge lengths in all the frames, in all the videos. For each performing person $i$, we compute the average length $\norm{e_{i}}$ of each edge $e = (p, q) \in E$, over all the video frames, to get a reliable measure of its length. We then for each edge $e = (p, q) \in E$, compute its average length $\norm{e}$ over all the performing people, to get the designated skeleton edge lengths. We then scale the edges in all the frames in all the videos, so their lengths will be equal to the designated skeleton. We use the Breadth-First Search (BFS) strategy to scan and alter the skeleton joints hierarchically. We start from an arbitrary joint $p$ and alter the joint locations according to their BFS order that starts from $p$. We set the new location of every neighboring joint $q$, by adding the vector $\overrightarrow{pq}$ to the new location of $p$, after scaling the vector to have the same magnitude as in the designated skeleton.

\begin{figure}[]
\centering
  \includegraphics[width=0.63\linewidth,keepaspectratio]{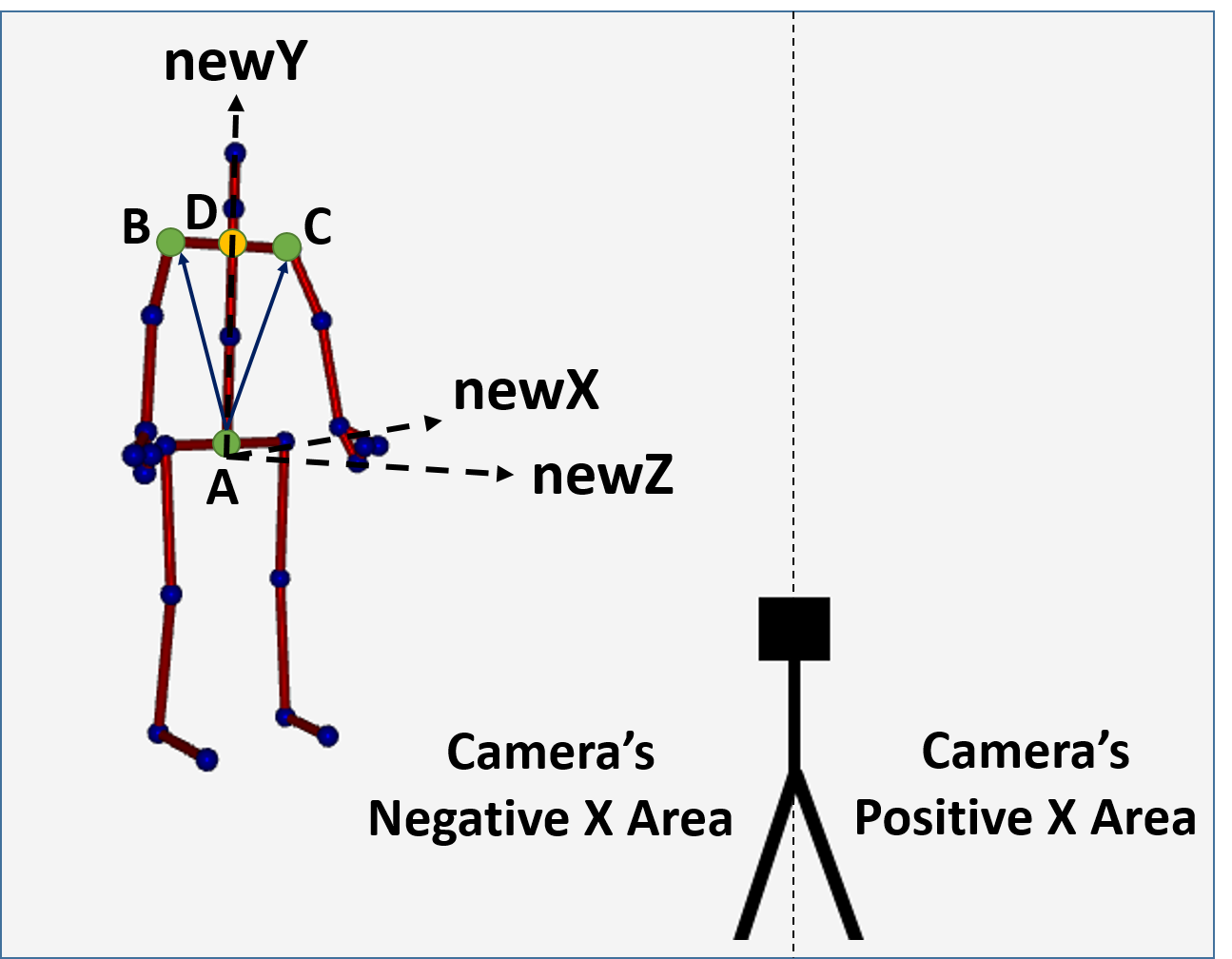}
  \caption[Skeleton body-center projection coordinate system]{Joints $A, B, C, D$ are Spine-Base, Shoulder-Left, Shoulder-Right and the average of B and C, respectively.}
  \label{fig:BodyCenterPlane}
\end{figure}

\paragraph{Alignment to Body-Plane Coordinate System.}
\label{lbl:skeleton-alignment}
We project all the joint 3D locations on a coordinate system that is based on a body-center plane that we compute at each frame. We use the locations of the Spine-Base, Shoulder-Left and Shoulder-Right joints, which we mark as $A, B ,C$, respectively, as visualized in Figure~\ref{fig:BodyCenterPlane}. Let us define the point $D = \frac{1}{2}(B + C)$, as the average of the two shoulders $B,C$. The vector $\hat{Y}$ will be the normalized vector $\overrightarrow{AD}$. The vector $\hat{Z}$ will be the normalized vector $\overrightarrow{AC} \times \overrightarrow{AB}$. The vector $\hat{X}$, will be the normalized vector $\hat{Y} \times \hat{Z}$. We project the joints on the system by multiplying them by a matrix that consists of $\hat{X},\hat{Y}$ and $\hat{Z}$ as row vectors. We finally subtract the Spine-Base joint location from all the joints, to translate the skeleton to the origin.

\subsection{Video Temporal Alignment}
\label{lbl:vidTemporalAlignment}

We detect mutual temporal PoIs in the training videos and force them to occur at the same time in all the videos, in order to have more meaningful information at each frame. We do that by temporally scaling the sequences between them, such that they will have an equal length in all the videos. Our scaling is done using linear interpolations of joint locations.

\paragraph{Active Joint Detection.} We want to discriminate between the active and inactive joints of the movement in order to detect the rest sequences of the active joints and use their temporal boundaries as temporal PoIs. Let us introduce $\overline{L}_{ji}$, as the mean of the joint locations $L_{jfi}$ of joint $j$ in video $i$, over all frames $f \in F_{i}$. We use it to compute the variance $\widehat{\sigma}^{2}(L_{ji})$:

\begin{equation}\label{eq:jointLocationVariance}
\widehat{\sigma}^{2}(L_{ji}) = \frac{1}{\norm{F_{i}} - 1}\sum_{f=1}^{\norm{F_{i}}}{\norm{L_{jfi} - \overline{L}_{ji}}^{2}} .
\end{equation}

We use the mean of the joint variances as a threshold between the inactive and active joints of the video. We eventually consider a joint as active in the movement, if at least $p \in [0,1]$ of the training videos agreed on it as such, where $p$ is a parameter. We used $p = 0.8$ in our experiments.

\begin{figure}[]
\centering
  \includegraphics[width=\linewidth,keepaspectratio]{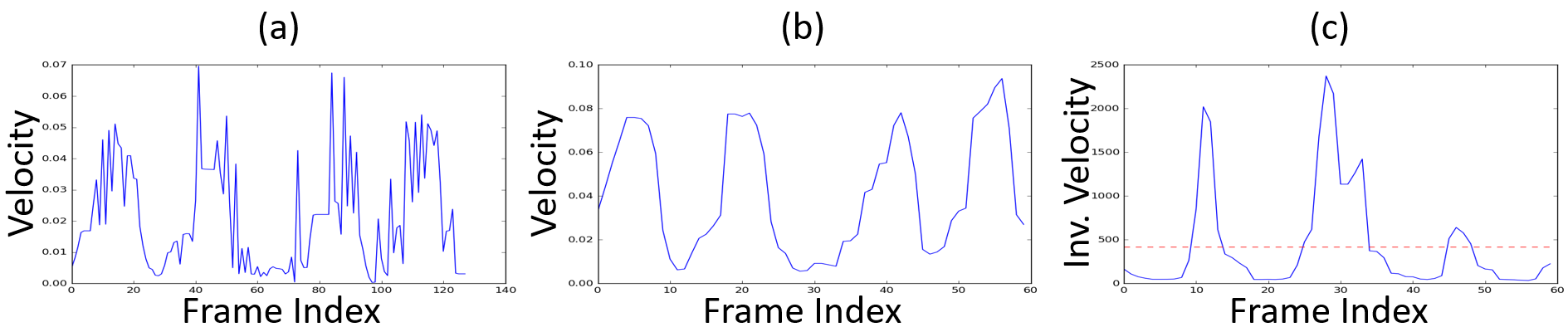}
  \caption[Rest sequences detection]{Detection of three movement-structural rest sequences using the velocities of the right hand. From left to right, the raw velocity magnitudes~(a) are time-filtered~(b) and finally flipped and thresholded~(c).}
  \label{fig:FindRests}
\end{figure}

\paragraph{Rest Sequences Detection.}
\label{lbl:rest-detection}
We aim to detect frame sequences in which the velocities of the active joints are close to $0$. We start by filtering and smoothing the noisy videos output by the device using a median filter and a temporal Gaussian pyramid. We then compute the joint velocities, given as the discrete temporal derivative of the video, which is the subtraction of the location vectors of each joint, for every pair of consecutive frames. We then apply a median filter on the velocities. We then compute $v^\eta$ as a function of the velocity $v$, which flips the velocity values, where $\eta < -1$ is a parameter. We use the mean as a threshold between rest and non-rest frames. We use $\eta = -1.5$ in our experiments. Figure~\ref{fig:FindRests} visualizes this process. We finally consider a frame as a rest frame, if at least $p \in [0,1]$ of the active joints consider it as such. In our experiments, we use $p = 0.3$. In order to turn the rest indices into rest sequences, we iterate on the indices. If the difference between the current index and the previous index is more than $\rho \in [0,1]$ of the video length, then we decide that the current index starts a new sequence. We use $\rho = 0.075$ in our experiments. We repeat this step over all the training videos, minding that different amounts of rest sequences can be detected for each video. We use the median over all the training videos in order to define the designated number of rests in the movement. For videos that reported more sequences, we only keep those that best match in time to the other videos. For videos that reported less sequences, we repeat the detection process, each time multiplying $\eta$ by $0.9$, until enough rest sequences are detected.

\begin{figure}[]
\centering
  \includegraphics[width=\linewidth,keepaspectratio]{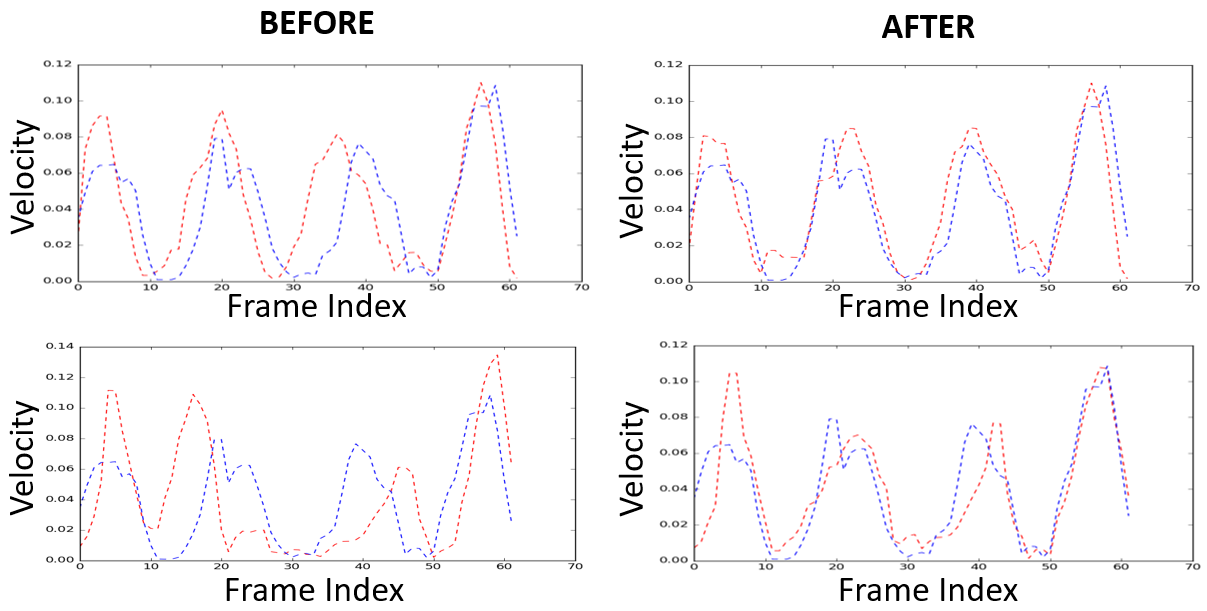}
  \caption[Rest alignment algorithm effectiveness]{The temporal PoI based warping algorithm in two running examples. The blue and red lines are the hand velocity magnitudes in the reference and aligned videos, respectively (best viewed in color).}
  \label{fig:RestAlignmentResults}
\end{figure}

\paragraph{Warping the Videos to Match Temporal PoIs.}
\label{lbl:warping-rests}
We use the rest sequences start and end frame indices as mutual PoIs and force them to occur at the same time in all the videos. We choose one of the videos to be a reference, such that all the other videos will be temporally aligned to it. The video that is closest to the centroid of all the videos in terms of PoI indices is selected. Then, for each video we temporally scale the sequences between every pair of adjacent PoIs, such that their length will be equal to those from the reference video. Figure \ref{fig:RestAlignmentResults} illustrates the effectiveness of this step.

\subsection{Model Learning}
\label{lbl:modelCreation}
Based on the fact that the skeletons in all the videos are normalized and fixed to the body-center plane and that the videos are aligned in time, the features we extract on every frame are the 3D locations of the joints, their 3D velocities, their pairwise distances and the angles between their edges. Additionally, we use the original lengths of the movement and the sequences from Section~\ref{lbl:vidTemporalAlignment} as time-related parameters. Let us introduce $o_{i}$ the observation of the parameter $o$ in video $i$. We only use scalars as observations, which means that vectors, such as locations and velocities, are separated into their three components. Let us introduce $\overline{o}$ as the mean of all the observations $o_{i}$ and $\overline{o^{i}}$ as the mean of all the observations excluding $o_{i}$. We use the leave-one-out method to learn the legitimate deviation of each parameter $o$. We do that by computing all the differences $|{o_{i} - \overline{o^{i}}}|$ and then computing their mean $M_{o}$ and standard deviation $S_{o}$, which we will use for normalizing deviations when we assess input videos. Our final trained model contains the triplets $(\overline{o}, M_{o}, S_{o})$ for every parameter. In addition, it contains the designated skeleton dimensions from Section~\ref{lbl:videoNormalization}, the video lengths, the detected active joints, the designated indices of the PoIs and the original lengths of the sequences from Section~\ref{lbl:vidTemporalAlignment}.

\begin{figure}[]
\centering
  \includegraphics[width=\linewidth,keepaspectratio]{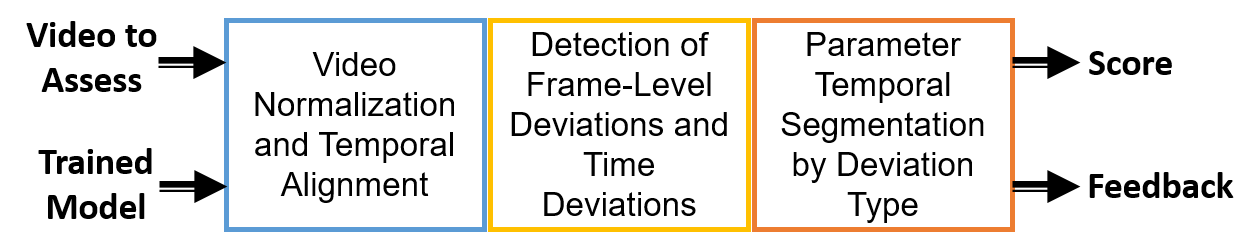}
  \caption[The Assessment Method in High Level]{The assessment method in high level.}
  \label{fig:AssessmentMethod}
\end{figure}

\section{Assessment Method}
\label{lbl:our-method-test}
As shown in Figure~\ref{fig:AssessmentMethod}, given a trained model and an input video to assess, we apply the normalization stage from Section~\ref{lbl:videoNormalization}, using the designated skeleton dimensions from the trained model. We then time-warp the input video using the algorithm from Section~\ref{lbl:vidTemporalAlignment}, using the detected active joints and rest sequences from the trained model. We then compute the parameter deviations. For each parameter $o$, we denote $d_{o}$ as the absolute difference between the input video value and the mean of observations $\overline{o}$, which we have computed in Section~\ref{lbl:modelCreation}. We then represent the difference in standard deviations, using the distance mean $M_{o}$ and standard deviation $S_{o}$, which we have in the model too:

\begin{equation}\label{eq:stdOfPartialDistances}
D_{o} = \frac{d_{o} - M_{o}}{S_{o} + \varepsilon} .
\end{equation}

We use $\varepsilon = 0.0005$ in our experiments, to avoid division by zero. We threshold the deviations in order to discriminate between accepted and unaccepted ones, using the threshold parameter $\gamma$, which we have set to $2.5$ standard deviations in our experiments:

\begin{equation}\label{eq:paramThreshold}
\begin{aligned}
D_{o} = \begin{cases}
D_{o} & D_{o} > \gamma \\ 
0 & otherwise

\end{cases}
\end{aligned}. 
\end{equation}

\subsection{Parameter Deviation Time Segmentation}
\label{lbl:paramDeviationTimeSegmentation}
\begin{figure}[]
\centering
  \includegraphics[width=\linewidth,keepaspectratio]{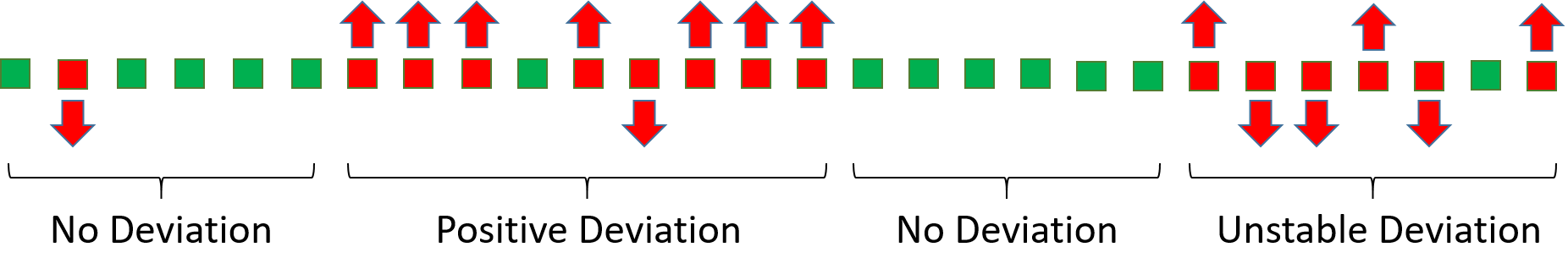}
  \caption[Parameter deviation time segmentation for feedback]{The segmentation aggregates the frame-level deviations (squares) of each scalar parameter into temporal segments, to produce higher-level, more meaningful feedback to the user. The algorithm aims to desirably temporally-segment the deviations, balancing between sequence classification significance and length (best viewed in color).}
  \label{fig:ParameterTimeSegmentation}
\end{figure}

We aim to aggregate the frame-level deviations into temporal segments that differ by deviation type, as illustrated in Figure~\ref{fig:ParameterTimeSegmentation}, which is essential for producing meaningful feedback and eliminate local deviation noise. As mentioned in Section~\ref{lbl:modelCreation}, all the deviations are scalars. Therefore, the aggregated deviation types we support are: 1. No Deviation. 2. Positive Deviation. 3. Negative deviation. 4. Unstable deviation.

\paragraph{Sequence Classification.}
\label{sec:seqClassification}
We build a classifier that will classify input sequences into one of the four deviation categories and output classification confidences, which will be used for finding the optimal time segmentation for each parameter. For that reason, let us introduce the rate measure $R_{T}$, which is the rate of the sequence indices that have deviation of type $T$. We then introduce the scattering measure $S_{T}$, which aims to tell how well deviation type indices are scattered in the sequence. It computes the index variance of deviation type indices, which we mark as $\widehat{\sigma}^{2}(seq_{T})$ and compares it to the index variance of all the sequence indices, which we mark as $\widehat{\sigma}^{2}(seq)$. A small difference between the variances will indicate a good scattering.

\begin{equation}\label{eq:feedbackScatteringMeasure}
S_{T} = \left(\frac{\widehat{\sigma}^{2}(seq_{T}) - \widehat{\sigma}^{2}(seq)}{\widehat{\sigma}^{2}(seq)}\right)^2 .
\end{equation}

We compute the classification score for each elementary deviation type $T$, which can be positive, negative or none, where $\lambda$ is a parameter that we set to $0.25$ in our experiments:

\begin{equation}\label{eq:feedbackElementaryClassScore}
Score_{T} = \lambda R_{T}S_{T} .
\end{equation}

We then compute the rate measure of the unstable deviation type $R_{\sim} = \min(1, \rho \cdot \min(R_{+}, R_{-}))$, where $\rho$ is a parameter that we set to $2$ in our experiments and $R_{+},R{-}$ are the rate measures of the positive and negative deviation types, respectively. The scattering measure $S_{\sim}$ is simply the average of the scattering measures of the positive and negative deviation types. The final score of the unstable deviation class is then computed as $Score_{\sim} = R_{\sim}S_{\sim}$. We finally add another dummy deviation type that is used to catch all the cases that do not have a clear classification, with score $Score_{\ast} = \min(1 - S_{+}, 1 - S_{-}, 1 - S_{0})$. We apply the Softmax function on the vector of scores and output the class with the highest score. If the dummy class has the highest score, then we output zero score, to prevent the sequence from becoming a segment.

\paragraph{Optimal Time Segmentation.}
The trivial, yet useless optimal segmentation will be treating each frame as a perfectly classified segment. We exclude such segmentations, by rewarding longer sequences, by multiplying their classification scores by $(1 + \ln(|seq|))^\xi$, where $|seq|$ is the sequence length and $\xi \in \mathbb{R}$ is a parameter. We use $\xi = 0.2$ in our experiments. Let us define $L = \{0, 1, ..., |seq| - 1\}$ as the set of all frame indices in the sequence $seq$. Let us denote $\mathbb{P}(L)$ as the power set of $L$, such that $|\mathbb{P}(L)| = 2^{|seq|}$ and such that every $l \in \mathbb{P}(L)$ is a sorted array. We want to find the subset of sorted indices $l \in \mathbb{P}(L)$ that gives the highest average classification score when used as boundaries, which denote $|l| + 1$ segments. We therefore want to find

\begin{equation}\label{eq:dpSegObjective}
\begin{aligned}
&\arg\!\max_{l \in \mathbb{P}(L)} \{ \frac{1}{|l| + 1}[cs(seq[0 \! : \! l[0]]) + \\
& \quad \quad \sum_{i=1}^{|l| - 1}{cs(seq[l[i - 1] \! : \! l[i]])} + cs(seq[l[|l| - 1] \! : \! |seq|])
\end{aligned} ,
\end{equation}

where the function $cs(sequence)$ returns the sequence classification score, multiplied by the sequence length reward and where $seq[i:j]$ is the subsequence of $seq$ from index $i$ to index $j$, non inclusive. In order to find the optimal segmentation, we implement a dynamic programming algorithm, with the array $Scores$ that will store for each index in the sequence the score of the best segmentation up to that index, such that for each $i \in \{1, 2, |seq| - 1\}$:

\begin{equation}\label{eq:feedbackSegmentStep}
\begin{aligned}
&Scores[i] = \max_{0 \leq j < i} \{ \frac{1}{Seg_{j} + 1} (Scores[j] \cdot Seg_{j} \\
&\quad \quad + cs(seq[j+1:i+1]))\}
\end{aligned} ,
\end{equation}

where $Seg_{j}$ denotes the number of segments in the optimal segmentation up to index $j$.

\subsection{Numerical Assessment Score}
\label{ref:produceScore}
We aggregate the parameter deviations into a final quality score. We reduce noise by utilizing the classified segments from Section~\ref{lbl:paramDeviationTimeSegmentation}, discarding all the deviations that were not a part of a segment classified as deviating. We divide our parameters into three sets: $A, N$ and $T$, which are the set of active-joint related parameters, the set of non-active joint related parameters and the set of time-related parameters, respectively, as described in Section~\ref{lbl:modelCreation}. For each set $S \in \{A, N, T\}$, we define the subset $S_{d} \subseteq S$ to be the set of all the deviating parameters from $S$. We reduce more noise by multiplying all the active-joint related deviations by $\alpha ^ d$, where $\alpha \in (0,1)$ is a parameter and $d$ is the average number of deviating segments. We do it as we expect longer, yet fewer deviating segments on improper performances. We use $\alpha = 0.75$ in our experiments. The score of the parameter set $S$ will be:

\begin{equation}\label{eq:paramSetScore}
Score(S) = 1 - \min(1, \frac{1}{|S|}\sum_{s \in S_{d}}{s}) ,
\end{equation}

We then compute the final score using the weight parameters  $\alpha_{A}, \alpha_{N}, \alpha_{T} \in \mathbb{R}$, such that $\alpha_{A} + \alpha_{N} + \alpha_{T} = 1$, which determine the weights of active joint related parameters, non-active joint related parameters and time parameters, respectively. The final score we output is:

\begin{equation}\label{eq:finalScore}
Score = \alpha_{A} Score(A) + \alpha_{N} Score(N) + \alpha_{T} Score(T) .
\end{equation}

In Section~\ref{lbl:results}, we demonstrate how the produced score accurately discriminates between movements of different performance qualities. We use $\alpha_{A} = 0.73$, $\alpha_{T} = 0.25$ and $\alpha_{N} = 0.02$ in our experiments. We use a low value for $\alpha_{N}$ in order to almost cancel its contribution to the score, while keeping the non-active joint deviations usable for textual feedback production.

\subsection{Feedback Generation}
\label{lbl:feedback} 
We utilize the output of the deviation segmentation algorithm described in Section~\ref{lbl:paramDeviationTimeSegmentation} and collect all the segments that were not classified as proper ones, where each segment represents the behavior of a single parameter over a range of time. Since all our parameters are scalars, the deviation types can only be too high, too low or unstable. An example for such a segment can be a too high horizontal velocity of the right elbow, between two points in time. In addition to the joint-related deviations, we also use all the time parameters. We compute the loss each segment has caused to the final score, according to the process described in Section~\ref{ref:produceScore} and sort them by importance, to output the most crucial and effective feedback items. We stop generating feedback items when the loss of the current item is less than a half of the loss of the previous item, or when we have already output five items. Figure~\ref{fig:feedbacks} demonstrates feedbacks that were generated by the algorithm.

\section{Experimental Results}
\label{lbl:results}

\begin{figure}[]
\centering
  \includegraphics[width=0.8\linewidth,keepaspectratio]{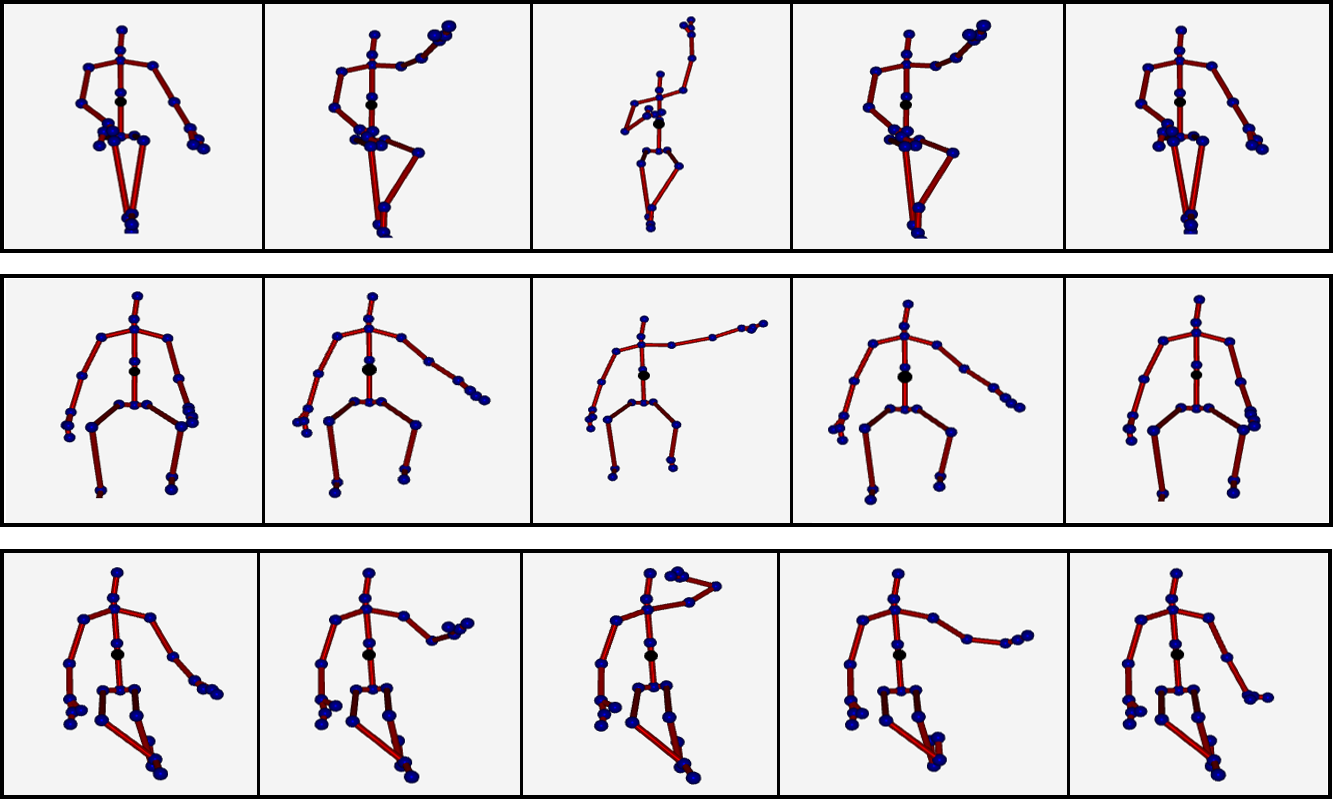}
  \caption[The tested FMA movements]{Representative frames from proper performances of the three FMA movement types our model has been trained and tested on. The first movement contains three structural rests, while the second and third movements contain one rest each.}
  \label{fig:motionRepFrames}
\end{figure}

\paragraph{FMA Movement Dataset.} We tested our method on a dataset of three FMA movement types that are illustrated in Figure~\ref{fig:motionRepFrames}. We acquired it in a hospital, under guidance of an occupational therapist, who also labeled the movements with FMA scores. These movements have 3 possible scores: 0 if the movement cannot be performed, 1 if it is improperly performed, or 2 if the movement is properly performed. The properly-performed movements, which are the only movements we use for training, were performed by the medical staff, men and women ranging between 30-60 years old. The other movements, were performed by the occupational-therapy department patients, men and women ranging between 45-75 years old, who were recovering after strokes. The numbers of training and testing video clips are detailed in Table~\ref{tbl:trainingVideos} and Table~\ref{tbl:testVideos}, respectively.

\paragraph{Movement Structures and Objectives.} As illustrated in Figure~\ref{fig:motionRepFrames}, in movement number 1, the patient should raise their hand in front of their body with a fully extended elbow, such that their shoulder is in 90\degsymb, then raise their hand until it is straight above their head, keeping their elbow extended, then lower their hand back to the front of the body and finally, lower and relax their hand. In movement number 2, the patient should raise their hand aside of the body with a fully extended elbow, such that their shoulder is in 90\degsymb, then lower and relax their hand. In movement number 3, the patient should raise their hand and touch their head, by raising their elbow aside of the body and bending it, with shoulder in 90\degsymb, then lower and relax their hand.

\begin{figure}[]
\centering
  \includegraphics[width=0.9\linewidth,keepaspectratio]{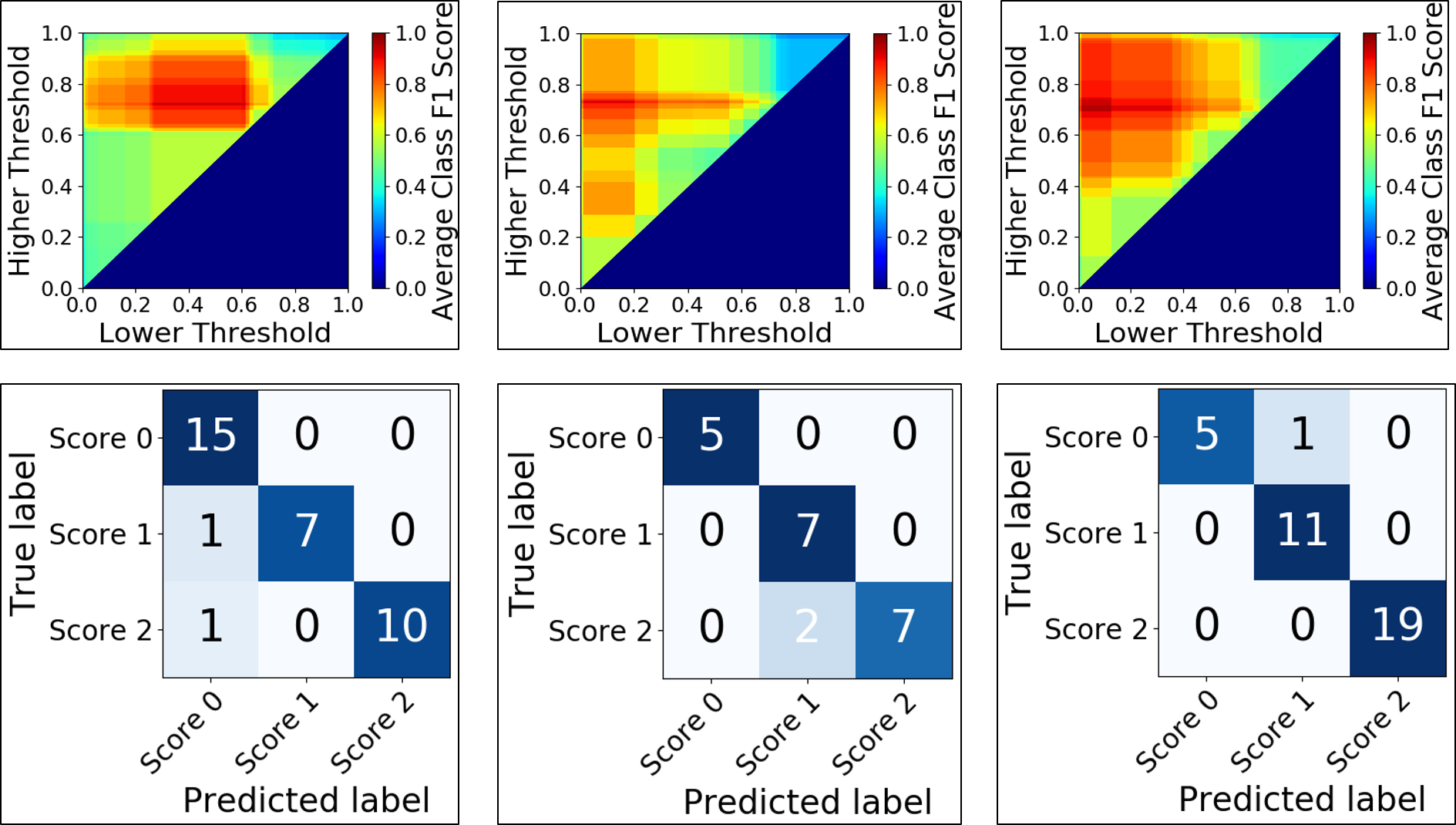}
  \caption[Classification performances of FMA movements]{From left to right, the average classification F1 scores at each pair of thresholds, of our three tested FMA movement types and their corresponding confusion matrices with the best pair of thresholds (best viewed in color).}
  \label{fig:classificationResults}
\end{figure}

\paragraph{Evaluation and Ablation Study.} We aim to show that even though the model has only been trained on properly-performed movements, the numerical scores it produces can accurately discriminate between videos from the three FMA score groups. When we assess videos from classes 0 and 1, we will use a model that has been trained on all the proper videos of the type of movement. When we assess videos from class 2, which are the proper videos, we will use a dedicated trained model for each tested video, such that the videos of the tested person will not be included in the training set. As illustrated in Figure~\ref{fig:classificationResults}, for each of the three movements, we look for the pair of thresholds that will most accurately divide the scores into the three classification groups and show the confusion matrix they produce. Table \ref{tbl:performances} shows the results of the ablation study we have conducted. Since our test data contains videos of real patients, who usually perform the movements slowly, we also show the classification accuracies with $\alpha_{T}=0$, to demonstrate that our model can accurately discriminate between the categories even when timings are ignored. In addition, we turn on and off the joint-grouping feature, denoted by \textbf{JG}, the time-warping feature, denoted by \textbf{TW} and the deviation-segmentation feature, denoted by \textbf{DS}, to demonstrate the importance of each. The joint grouping feature is responsible for giving joints that were detected as active a higher weight in the score. The time-warping feature is responsible for video alignment during training and testing. The deviation-segmentation feature is responsible for exclusion of deviating parameters that do not belong to a deviating segment. The ablation study demonstrates the essentiality of each component and the superiority of our temporal PoI based time warping algorithm over DTW. The COW time-warping results were excluded, as they underperformed both DTW and our warping algorithm.

\begin{table}[]
\centering
\resizebox{0.65\linewidth}{!}{%
\begin{tabular}{l|c|c|c}
\hline
\hline
\textbf{Movement} & \textbf{\#Clips} & \textbf{\#People} & \textbf{\#People with Two Clips} \\ \hline
\textbf{Movement 1}            & 11              & 7              & 4             \\
\textbf{Movement 2}            & 9               & 5              & 4             \\
\textbf{Movement 3}            & 19              & 10             & 9             \\
\hline
\hline
\end{tabular}
}
\caption[Training videos of properly performed movement]{The numbers of training videos and performing people per movement type. The training videos only contain properly-performed movements.}
\label{tbl:trainingVideos}
\end{table}

\begin{table}[]
\centering
\resizebox{0.52\linewidth}{!}{%
\begin{tabular}{l|c|c}
\hline
\hline
\textbf{Movement} & \textbf{\#Class 0 clips} & \textbf{\#Class 1 clips} \\ \hline
\textbf{Movement 1}            & 15              & 8*              \\
\textbf{Movement 2}            & 5               & 7               \\
\textbf{Movement 3}            & 6               & 11               \\
\hline
\hline
\end{tabular}
}
\caption[Number of testing videos]{The number of testing videos of classes $0, 1$ per movement. (*) As we only had a single genuine video of class 1 for Movement 1, we added seven artificial videos, by applying minor distortions on proper videos.}
\label{tbl:testVideos}
\end{table}

\begin{table*}[]
\centering
\resizebox{0.71\linewidth}{!}{%
\begin{tabular}{l|c|c|c|c|c}
\hline
\hline
\textbf{Setting}   & \textbf{Metric}     & \textbf{Movement 1} & \textbf{Movement 2} & \textbf{Movement 3} & \textbf{Average} \\
                   &                     & $\alpha_{T} = 0.25$ / $\alpha_{T} = 0$ & $\alpha_{T} = 0.25$ / $\alpha_{T} = 0$ & $\alpha_{T} = 0.25$ / $\alpha_{T} = 0$ & $\alpha_{T} = 0.25$ / $\alpha_{T} = 0$ \\ \hline
\textbf{TW}        & Mean F1       & 0.73 / 0.72  & 0.80 / 0.57 & 0.78 / 0.70 &  0.77 / 0.66    \\
                   & MSE           & 0.32 / 0.41  & 0.33 / 1.00 & 0.17 / 0.25 &  0.27 / 0.55    \\ \hline
\textbf{JG+TW}     & Mean F1       & 0.76 / 0.83  & 0.78 / 0.58 & 0.80 / 0.73 &  0.78 / 0.71    \\
                   & MSE           & 0.29 / 0.24  & 0.24 / 0.43 & 0.14 / 0.22 &  0.22 / 0.30    \\ \hline
\textbf{JG}        & Mean F1       & 0.77 / 0.82  & 0.86 / 0.70 & 0.78 / 0.76 &  0.80 / 0.76    \\
                   & MSE           & 0.59 / 0.18  & 0.14 / 0.24 & 0.17 / 0.19 &  0.30 / 0.20    \\ \hline 
\textbf{DS}        & Mean F1       & 0.83 / 0.82  & 0.82 / 0.78 & 0.85 / 0.84 &  0.83 / 0.81    \\
                   & MSE           & 0.24 / 0.62  & 0.19 / 0.24 & 0.14 / 0.14 &  0.19 / 0.33    \\ \hline
\textbf{DS+TW}     & Mean F1       & 0.84 / 0.74  & 0.82 / 0.77 & 0.85 / 0.82 &  0.84 / 0.78    \\
                   & MSE           & 0.32 / 0.68  & 0.19 / 0.24 & 0.14 / 0.17 &  0.22 / 0.36    \\ \hline
\textbf{DS+JG}     & Mean F1       & 0.93 / 0.90  & 0.87 / 0.81 & 0.91 / 0.87 &  0.90 / 0.86    \\
                   & MSE           & 0.06 / 0.18  & 0.14 / 0.19 & 0.06 / 0.11 &  0.09 / 0.16    \\ \hline
\textbf{DS+JG+DTW} & Mean F1       & 0.96 / 0.90  & 0.83 / 0.74 & 0.94 / 0.91 &  0.91 / 0.85      \\
                   & MSE           & 0.03 / 0.18  & 0.19 / 0.24 & 0.06 / 0.08 &  0.09 / 0.17    \\ \hline
\hline
\textbf{DS+JG+TW}  & Mean F1       & 0.94 / 0.90  & 0.92 / 0.81 & 0.96 / 0.91 & \textbf{0.94} / \textbf{0.87}             \\
                   & MSE           & 0.15 / 0.18  & 0.10 / 0.19 & 0.03 / 0.08 & \textbf{0.09} / \textbf{0.15}   \\ 
\hline
\hline
\end{tabular}
}
\caption[Performances comparison on three tested FMA movements.]{Ablation study. The mean F1 score and the corresponding MSE refer to the classification with the best pair of thresholds.}
\label{tbl:performances}
\end{table*}

\paragraph{Produced Feedback Examples.}
\label{sec:feedbackExamples}
\begin{figure*}[]
\centering
  \includegraphics[width=0.75\linewidth]{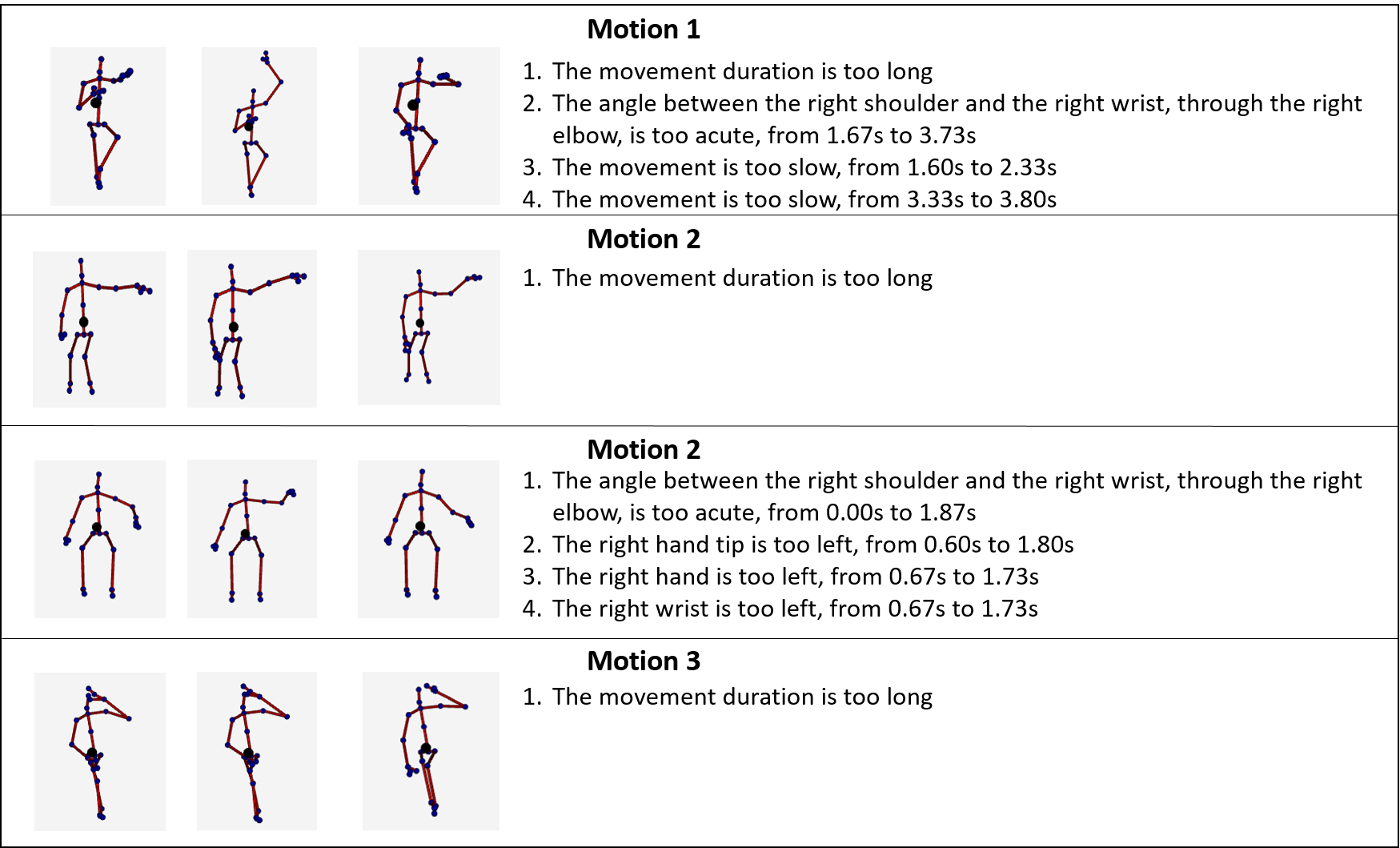}
  \caption[Examples of textual generated feedback on tested movements]{The generated textual feedbacks on improperly performed movements, using the trained models, which were only trained on proper movements.}
  \label{fig:feedbacks}
\end{figure*}
We used the feedback generation method detailed in Section~\ref{lbl:feedback}, to generate textual feedback for the most crucial deviations of each of the tested videos. As explained in the method in Section~\ref{lbl:feedback}, the number of feedback items can differ between videos and depends on their relative significance. Figure~\ref{fig:feedbacks} shows representative frames from improperly performed movements and the feedback that has been generated for them.

\section{Conclusions}
\label{lbl:conclusion}
In this work, we presented an automatic end-to-end movement assessment and feedback generation algorithm that only learned from properly-performed movement videos, without further annotations and which overcame the data noisiness of the relatively cheap Kinect2 device. We introduced a novel continuous time-warping algorithm, based on mutual PoIs that were automatically detected and extracted from the training samples. Using additional techniques, such as active joint detection and parameter deviation time-segmentation, we demonstrated the robustness of our model on FMA movements, even without time-warping and when ignoring the almost obviously deviating patient timings. 

In the future, we plan to extend this work and demonstrate its natural adaptation to other medical tests, such as Berg Balance Scale (BBS)~\cite{bbs}. Additionally, we plan to extend and improve the work by extending the variety of PoI types used as warping anchors. Another future direction may be the synthesis of training samples from our statistical model, which may serve as an input for learning algorithms that are based on the availability of large training sets.

\ificcvfinal
\paragraph{Acknowledgements.} We would like to thank the department of occupational therapy in the Galilee Medical Center for their cooperation and in particular Dorit Itah, who guided and labeled the FMA tests. We would also like to thank Nadav Eichler, Shmulik Raz and Hagit Hel-Or for their cooperation. Lastly, we would like to thank the CVPM'19 reviewers for their important insights.
\fi

{\small
\bibliographystyle{ieee}
\bibliography{citations}
}

\end{document}